\documentclass{bmvc2k}
\usepackage{marvosym}

\def\blue#1{\textcolor{blue}{#1}}

\def\blue#1{\textcolor{blue}{#1}}

\title{FastForensics: Efficient Two-Stream Design for Real-Time Image Manipulation Detection}

\addauthor{Yangxiang Zhang}{zhangyangxiang@stu.ouc.edu.cn}{1}
\addauthor{Yuezun Li}{liyuezun@ouc.edu.cn}{1,\blue{\text{\Letter}}}
\addauthor{Ao Luo}{aoluo@swjtu.edu.cn}{2}
\addauthor{Jiaran Zhou}{zhoujiaran@ouc.edu.cn}{1}
\addauthor{Junyu Dong}{dongjunyu@ouc.edu.cn}{1}

\addinstitution{
    School of Computer Science and Technology, \\ 
    Ocean University of China\\
}

\addinstitution{
 Southwest Jiaotong University \\
 \vspace{0.5cm}
  \blue{\text{\Letter}}: \footnotesize{Corresponding author}
}

\runninghead{Zhang et al}{FastForensics}

\def\eg{\emph{e.g}\bmvaOneDot}

\def\ie{\emph{i.e.}}

\usepackage{multirow}
\usepackage{amsmath} 
\usepackage{wrapfig}  
\usepackage{graphicx}
\usepackage{paralist}
\usepackage{amssymb}

\begin{document}

\maketitle

\begin{abstract}
With the rise in popularity of portable devices, the spread of falsified media on social platforms has become rampant. This necessitates the timely identification of authentic content. However, most advanced detection methods are computationally heavy, hindering their real-time application. In this paper, we describe an efficient two-stream architecture for real-time image manipulation detection. Our method consists of two-stream branches targeting the cognitive and inspective perspectives. In the cognitive branch, we propose efficient wavelet-guided Transformer blocks to capture the global manipulation traces related to frequency. This block contains an interactive wavelet-guided self-attention module that integrates wavelet transformation with efficient attention design, interacting with the knowledge from the inspective branch. The inspective branch consists of simple convolutions that capture fine-grained traces and interact bidirectionally with Transformer blocks to provide mutual support.
Our method is lightweight ($\sim$ 8M) but achieves competitive performance compared to many other counterparts, demonstrating its efficacy in image manipulation detection and its potential for portable integration.
\end{abstract}



\section{Introduction}
\label{sec:intro}
The rapid advancement of multimedia tools has greatly enhanced image manipulation techniques, making them more imperceptible and effortless. The abuse of these techniques can create plenty of misinformation and fabrication, raising serious concerns for social trustfulness, such as fake news, economic fraud, and blackmail~\cite{zampoglou2017large}. In recent years, notable efforts have been made to detect image manipulations, \ie, localizing the manipulated regions~\cite{wu2019mantra,liu2022pscc,wang2022objectformer,hu2020span}. These methods are commonly developed based on strong base models such as advanced Convolutional Neural Networks (CNNs) and Transformers, due to their powerful learning ability demonstrated on various tasks. 

These methods have shown promising detection performance on many datasets, aligning with the current focus of the field on pursuing detection accuracy. However, as image manipulation techniques become more popular and social apps become more prevalent, accuracy alone is no longer the only criterion for detection methods. In practical scenarios where tremendous fabricated media is daily spread over social platforms (\eg, FaceBook, TikTok, etc), it is essential to timely inform users about the authenticity of the media. Existing methods are unlikely to achieve this goal due to their heavy computational requirements, hindering their potential for efficient execution on portable devices.  

\begin{wrapfigure}{r}{0.5\textwidth}
\centering
\includegraphics[width=\linewidth]{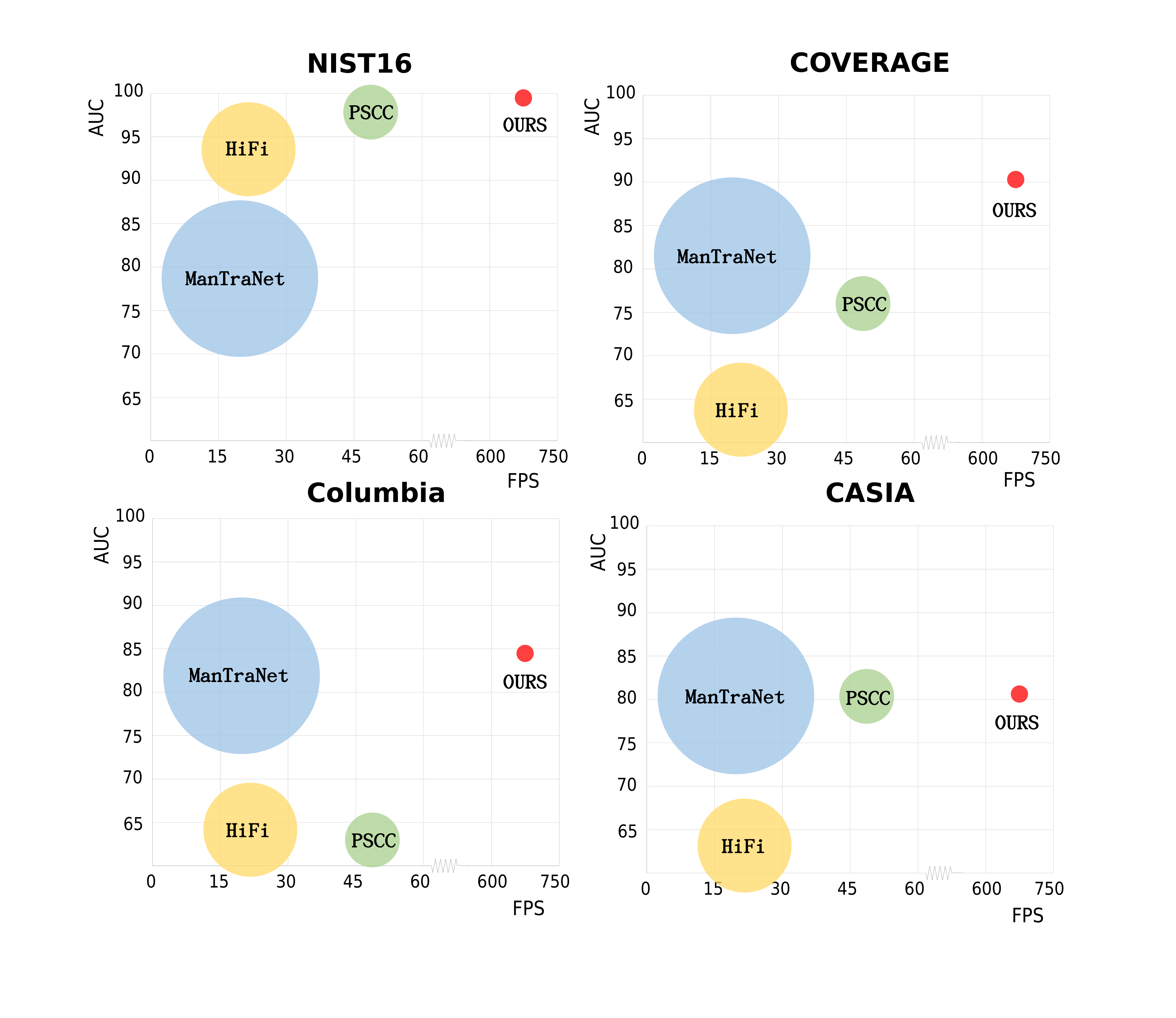} 
\vspace{-1.3cm}
\caption{\small AUC (\%), FLOPs (circle area), FPS of different methods. Our method (red circle) strikes a good balance between performance and efficiency.}
\label{fig:1}
\end{wrapfigure}

\begin{figure*}[!t]
\centering
\includegraphics[width=\linewidth]{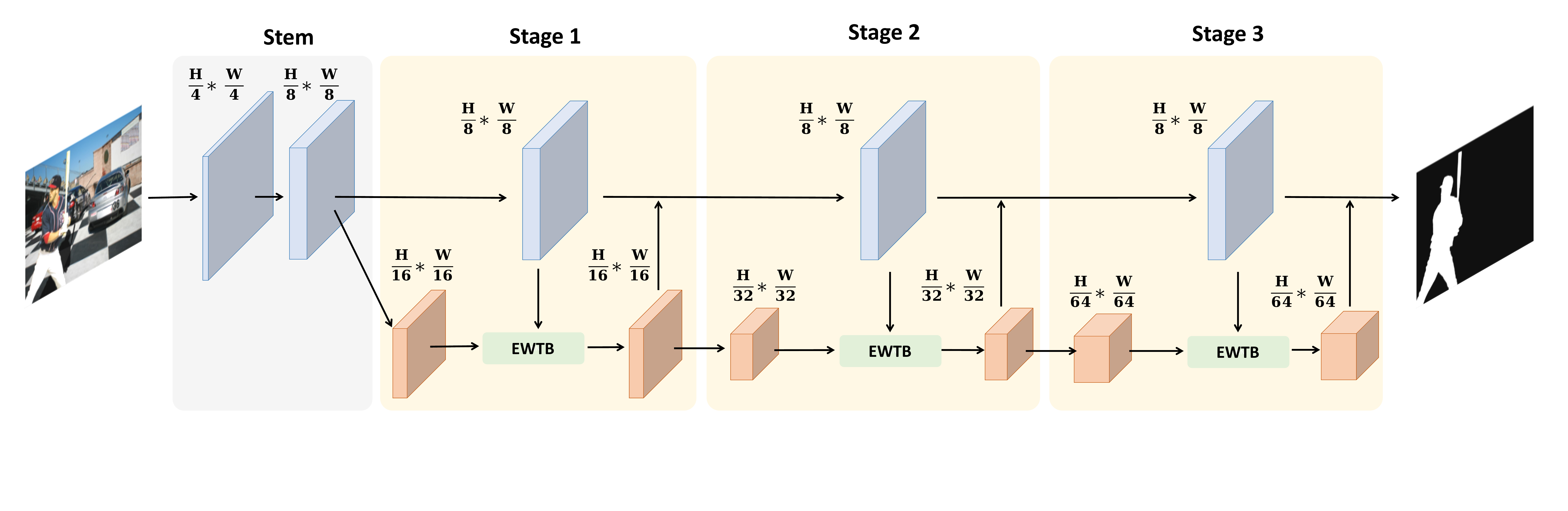}
\vspace{-1.5cm}
\caption{\small  {Overview of the architecture of our method.} \small The blue and orange cuboids represent the features from the inspective branch and the cognitive branch. EWTB is denoted as an Efficient Wavelet-guided Transformer Block.}
\label{fig:2}
\vspace{-0.5cm}
\end{figure*}

In this paper, we propose a new efficient two-stream design for real-time image manipulation detection, dubbed as {\em FastForensics}, which strikes a better trade-off between performance and efficiency (see Fig.~\ref{fig:1}). Our method is motivated by the detection process of human brains, which first understands the context and roughly spots the manipulated regions (cognitive view), then looks into the details of manipulated regions (inspective view). To imitate this process, we design two-stream branches corresponding to these two views (see Fig.~\ref{fig:2}). In the cognitive branch, we propose Efficient Wavelet-guided Transformer Blocks (EWTB) to capture the global relationships. Compared to the conventional Transformer block, the proposed block is greatly simplified using feature splitting and shared query and value features (see Fig.~\ref{fig:3}). Inspired by the effect of frequency in exposing manipulation traces~\cite{sun2023safl, wang2022objectformer}, we develop a new Interactive Wavelet-guided Self-Attention (IWSA) module to concentrate more on manipulation traces. This module integrates wavelet transformation into Transformer blocks, with the interaction with the knowledge from the inspective branch (see Fig.~\ref{Fig:IWSA}). The inspective branch is constructed using several depth-wise convolution layers. Due to the local receptive field in convolutions, this branch naturally focuses more on the fine-grained manipulation traces. To mutually boost these two branches, we integrate the intermediate features from the cognitive branch to the inspective branch, to provide more global instruction. Furthermore, the fine-grained knowledge from the inspective branch is then employed to instruct the learning of ETWB, enhancing the self-attention performance in the cognitive branch.
Our method is validated on several public datasets and compared to various state-of-the-art counterparts, showing the efficacy and efficiency of our method in detecting image manipulations.

The contribution of this paper is summarised as follows:
\begin{compactenum}
\item Given the significance of timely detection in the task of image manipulation, we describe a novel method called {\em FastForensics}, for striking a better trade-off between performance and efficiency. To the best of our method, this field remains largely unexplored, making our method a pioneering contribution in this field.

\item We propose a new Efficient Wavelet-guided Transformer Block (EWTB) that can integrate the wavelet transformation with an efficient and interactive design to highlight the manipulation traces. 

\item Our method is validated on standard datasets and compared with many state-of-the-art counterparts, showing its efficacy and efficiency performance. We also thoroughly study the effect of each module, {filling the gap of limited experimental experience in this field.}
\end{compactenum}

\section{Related Works}
\noindent{\bf Image Manipulation Detection}  
has become increasingly challenging with the advancement of
image editing tools and emerging AI-based techniques. To better detect the manipulations, current methods mainly employ sophisticated deep neural networks to learn and extract the traces of manipulation. These methods employ various backbone architectures, ranging from convolutional neural networks~\cite{wu2019mantra,liu2022pscc,zhou2023pre} to more recent Transformers~\cite{wang2022objectformer}. 
For instance, PSCCNet~\cite{liu2022pscc} extracts hierarchical features with a top-down path using HRNet and detects whether the input image has been manipulated using a bottom-up path. However, the intermediate process involves high-dimensional feature vectors, introducing more FLOPS. 
Objectformer~\cite{wang2022objectformer} combines RGB and frequency features to identify the tampering artifacts based on a Transformer architecture. Although it performs well in accuracy, the Transformer architecture has high computational complexity and resource consumption. Such high computational cost associated with these sophisticated models hinders their deployment on portable devices. 
Therefore, there is a pressing need to explore the efficiency of image manipulation detection methods.

\smallskip
\noindent{\bf Efficient Image Semantic Segmentation}  
is a task that aims to efficiently segment regions with different semantic categories, \ie, assigning semantic labels (such as cat, dog, etc) to each pixel~\cite{yu2018bisenet,cai2022efficientvit,xu2023pidnet,guo2022segnext,yu2021bisenet,xie2021segformer,hong2021deep}. 
This task is similar to our task, as it also assigns labels to pixels, but the difference is that the image manipulation detection task focuses on identifying whether a pixel is authentic or not. Thus both these tasks belong to pixel-wise classification tasks.
While it shares similarities with ours in terms of pixel-wise classification, these methods are not directly applicable to our task. This is because they are designed to capture the semantic differences (\eg, cat, dog), whereas our task focuses on extracting subtle manipulation traces that are independent of semantic information. The experiments in Sec.~\ref{results} have verified the limitations of directly applying the image semantic segmentation methods to our task. Therefore, in this paper, we propose a dedicated model with an efficient design specifically tailored for image manipulation detection.

\section{Method}
\subsection{Two-Stream Design}
Our method incorporates two branches: the cognitive branch and the inspective branch. The cognitive branch aims to comprehend the context at a glance and determine where the manipulated regions exist. To achieve this, we design Efficient Wavelet-guided Transformer Blocks (EWTB) that can efficiently capture global frequency-related relationships in manipulations. On the other hand, the inspective branch is composed of simple convolution layers that focus on local detail and extract the specific shape of the manipulated regions. To enhance this branch, we leverage the knowledge extracted from the cognitive branch as priors, which are fused with the inspective features. {Similarly, we improve the cognitive branch by incorporating knowledge from the inspective branch, guiding the learning of value features from our proposed Transformer blocks}. Then we employ various loss terms to supervise the training process. The overview of our method is shown in Fig.~\ref{fig:2}.


\begin{figure}[!t]
    \centering
    \begin{minipage}[t]{0.48\linewidth}
    \centering 
    \includegraphics[width=0.85\linewidth]{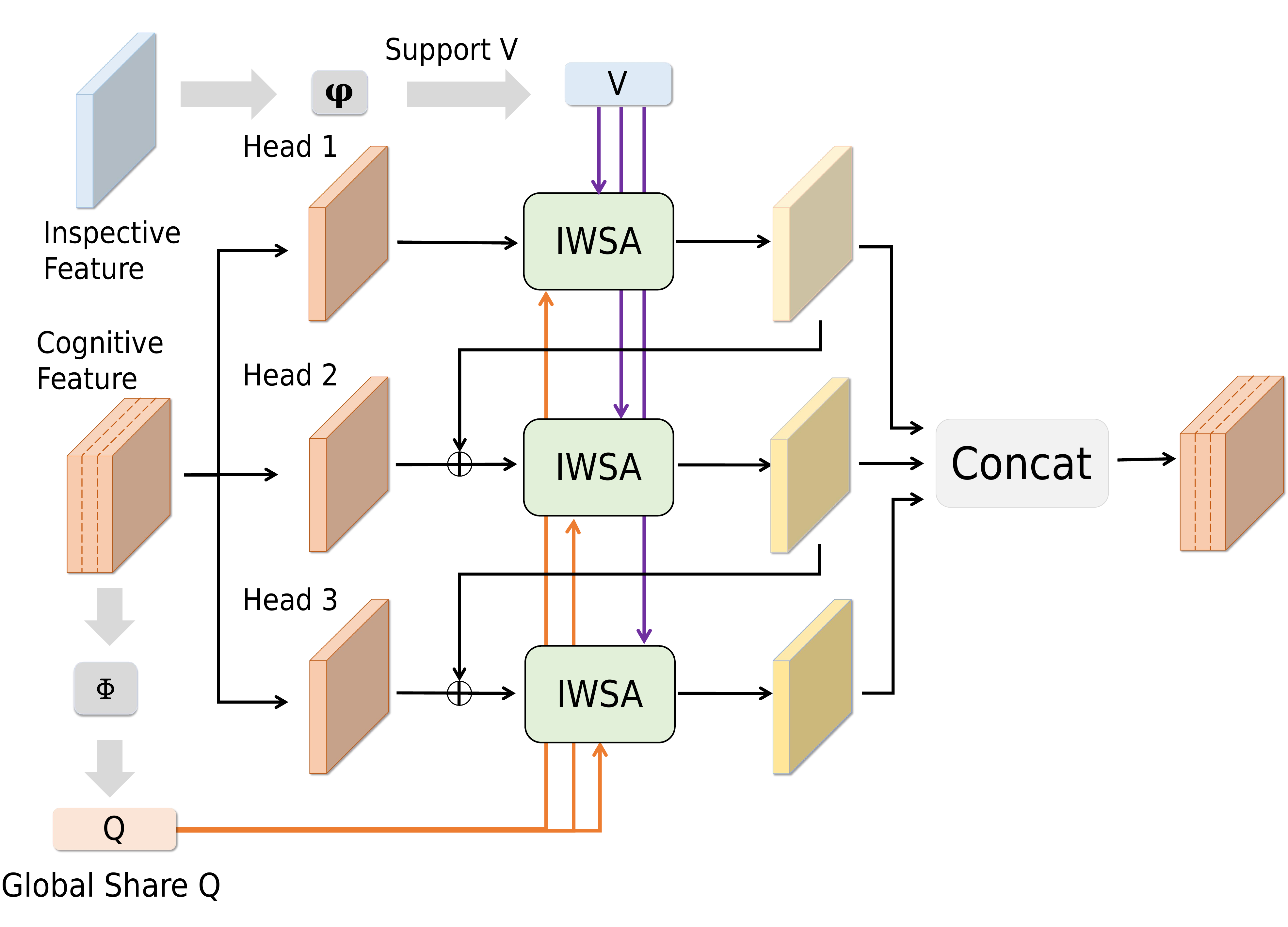}
    \vspace{-0.3cm}
    \caption{\small Overview of efficient wavelet-guided Transformer block (EWTB). } 
    \label{fig:3}
    \end{minipage}
    \hfill
    \begin{minipage}[t]{0.48\linewidth}
    \centering
    \includegraphics[width=0.85\linewidth]{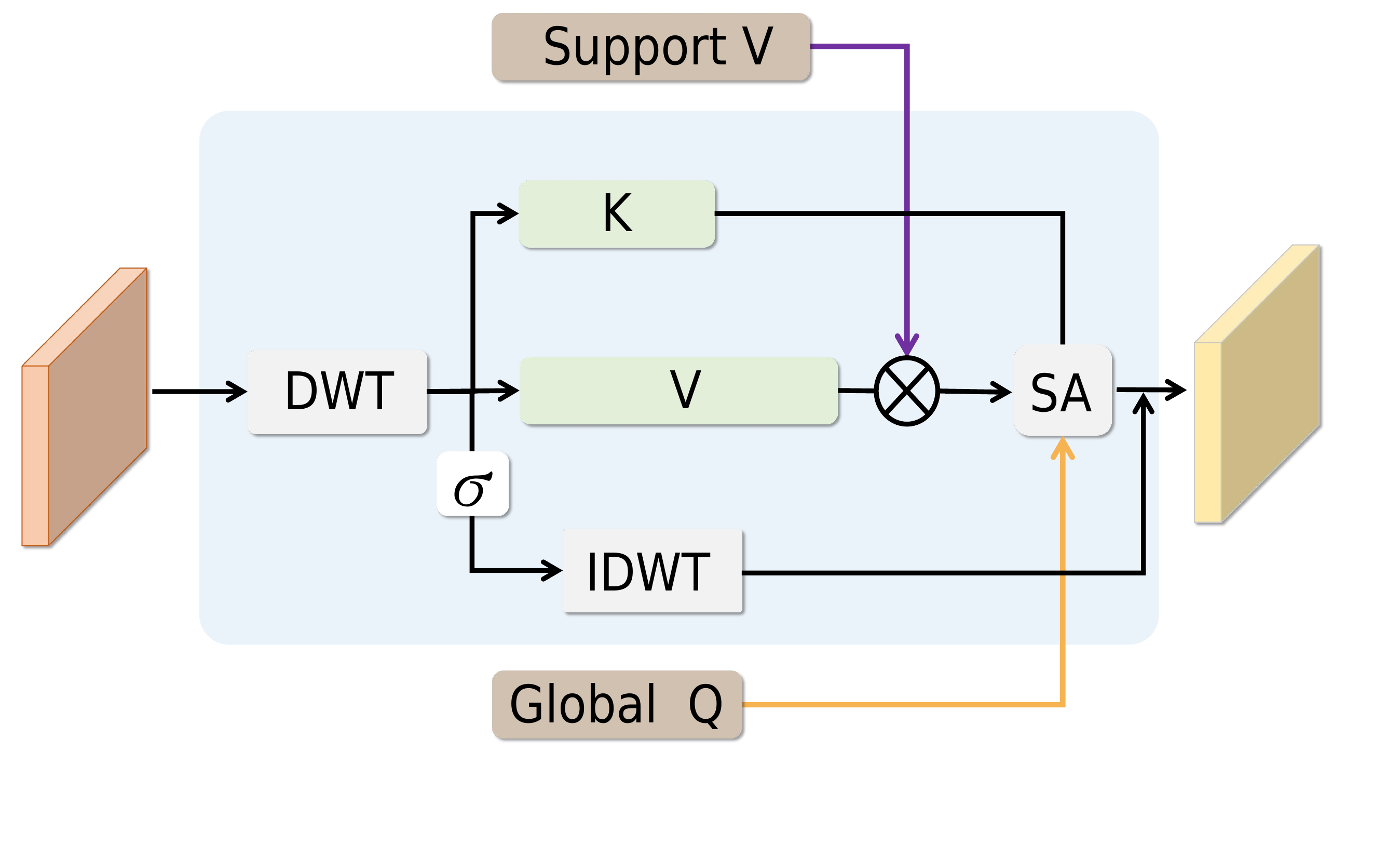}
    \vspace{-0.3cm}
    \caption{\small Overview of interactive wavelet-guided self-attention (IWSA). }
    \label{Fig:IWSA}
    \end{minipage}
\end{figure}

\subsection{Efficient Wavelet-guided Transformer Block}
Motivated by the effectiveness of self-attention~\cite{vaswani2017attention} in capturing global information, we adopt the principles of Transformer blocks in the cognitive branch. However, the conventional Transformer block requires large computational resources due to the $O(n^2)$ complexity of self-attention operations, which restricts its direct application to our task. {Moreover, the conventional Transformer blocks are designed for traditional vision tasks, prioritizing the extraction of semantic features rather than the subtle manipulation traces. } Hence, 
we propose a new Efficient Wavelet-guided Transformer Block (EWTB) dedicated to image manipulation detection (see Fig.~\ref{fig:3}).





\smallskip
\noindent{\bf Feature Decomposition.}
In vanilla self-attention~\cite{vaswani2017attention}, the major computational cost derives from the heavy linear transformations for query, key, and value features. Denote the input feature as $X$, the attention map $X'$ can be obtained by
\begin{equation}
\small
    \begin{aligned}
    H_i = {\rm Attn}(X W^{Q}_i, X W^{K}_i, X W^{V}_i), \;
    X' = {\rm Concat}(H_1,...,H_n) W^{O},
    \end{aligned}
    \label{eq:msa}
\end{equation}
where $W^{Q}_i,W^{K}_i,W^{V}_i$ are learnable parameters. 
To save the computation cost, simplifying the attention calculation is critical. Following~\cite{liu2023efficientvit,liu2018rethinking,yang2021nvit}, we decompose the input features into different pieces and perform self-attention only on the respective piece, as
\begin{equation}
\small
\begin{aligned}
H_i & = {\rm Attn}(x_i W^{Q}_i, x_i W^{K}_i, x_i W^{V}_i), \\
\end{aligned}
\label{eq:msa}
\end{equation}
where $X = [x_1, ..., x_n]$ and $x_i$ corresponds to the $i$-th feature piece for calculating the attention within $i$-th head.
This decomposition can greatly reduce the computation cost.

\smallskip
\noindent{\bf Interactive Wavelet-guided Self-Attention.} 
We then describe an Interactive Wavelet-guided Self-Attention (IWSA) module to capture the correlations in manipulation (see Fig.~\ref{Fig:IWSA}). 

{\em - Shared Global Queries}. While feature decomposition can greatly save computation costs, the different pieces are fully split, breaking their connections with each other. Thus the attention calculated on each piece can only consider the relationship inside, limiting their ability for global modeling. Considering that queries are important features that can provide sufficient indexing information, we formulate a shared global query for all heads, to better utilize the global information while further reducing the computation costs. Specifically, we generate the queries from the features before feature decomposition. This operation can be formulated as
\begin{equation}
\small
\begin{aligned}
Q = \phi(X), \;
H_i = {\rm Attn}(Q, x_i W^{K}_i, x_i W^{V}_i), \;
X' = {\rm Concat}(H_1,...,H_n) W^{O},
\end{aligned}
\label{eq:msa}
\end{equation}
where $\phi(\cdot)$ denotes a convolution operation to transform the input feature into queries.
To further improve the capacity, we use cascade connections~\cite{liu2023efficientvit} between adjacent heads to refine the attention maps as $x_{i+1} \leftarrow x_{i+1} + H_{i}$.

\begin{wrapfigure}{r}{0.5\textwidth}
    \centering
    \includegraphics[width=\linewidth]{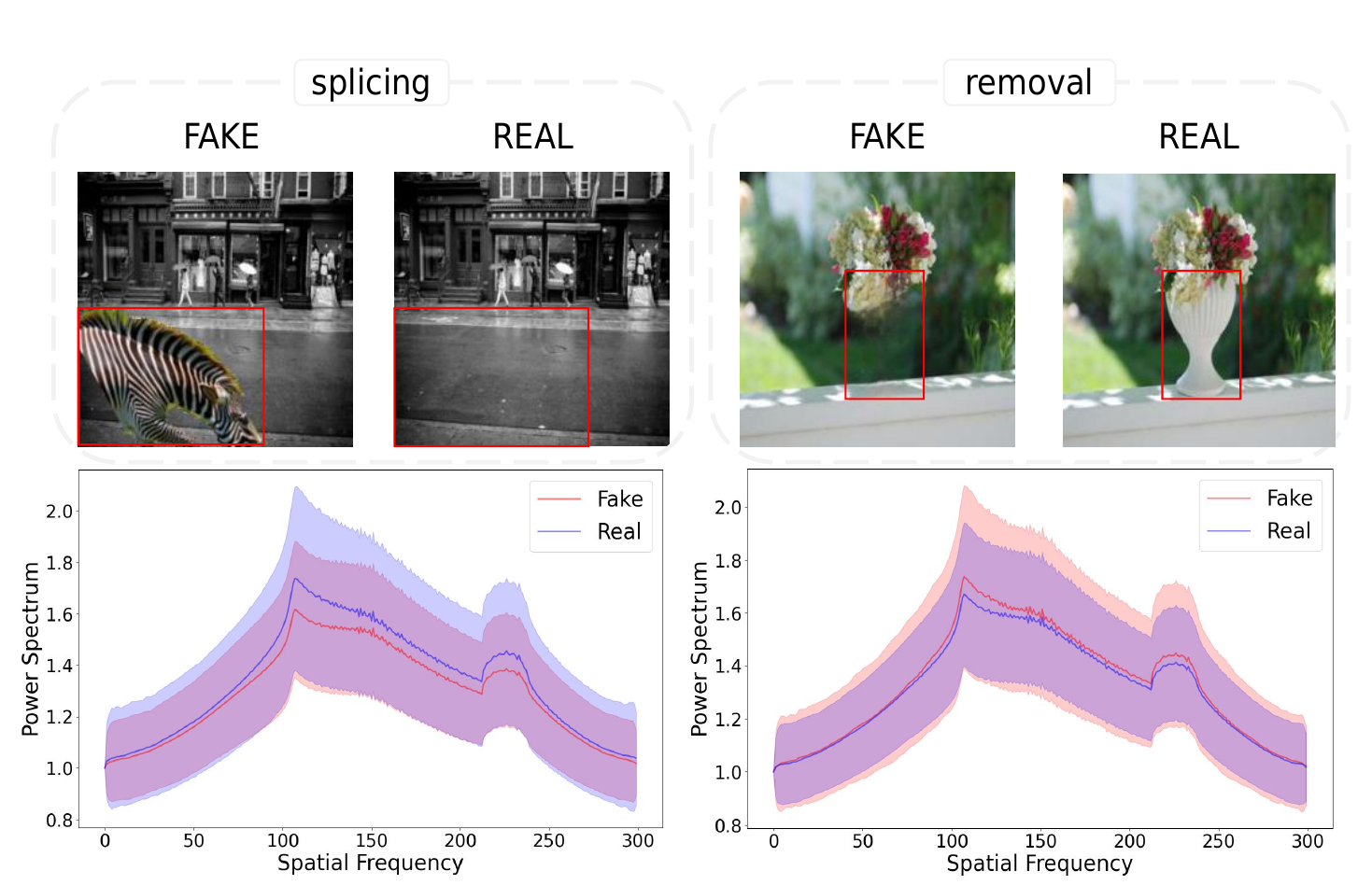} 
    \vspace{-0.6cm}
    \caption{\small Frequency statistics for real and fake images using splicing and object removal.}
    \label{fig:fre}
\end{wrapfigure}

{\em - Wavelet-guided Self-Attention.} 
Previous methods~\cite{zhou2018learning,wang2022objectformer} have utilized the frequency domain to uncover manipulation traces. However, these methods lack an in-depth analysis of whether manipulation traces can be reflected in the frequency domain. To verify this, we perform frequency analysis for real and fake images using two manipulation types of splicing and object removal in PSCC~\cite{liu2022pscc} dataset. {Using these two types is because we can find the image pairs of real and fake.} This analysis is performed on the manipulated region using its tight bounding box (red box in the top row) and the corresponding region in the real image. Fig.~\ref{fig:fre} shows the frequency difference exists between real and manipulated regions.

Note that Fourier Transforms such as DCT operations are widely used to extract frequency information. However, these operations have two main limitations: 1) they focus on the global frequency while neglecting local frequency details; 2) they fail to retain the position information in the transformed frequency domain, which is crucial in the context of attention modeling~\cite{liu2023forgery}
In contrast, wavelet transformation can analyze the local variation of different subspaces in the frequency domain while retaining the spatial structure of images~\cite{li2020wavelet,yao2022wave}. Inspired by this, we integrate Wavelet transformation into self-attention to highlight the importance of manipulation traces.


Specifically, the input feature can be transformed by Discrete Wavelet Transform (DWT) into four sub-band features, which are the arbitrary combinations of low-pass (L) and high-pass (H) filters. Thus we can denote the filtered features as LL, LH, HL, and HH respectively. Each feature focuses on a specific combination of low and high-frequency information. Then we concatenate these sub-band features as the input feature for self-attention. To compensate for the spatial information, we build a skip connection before and after the self-attention. In this connection, we first perform convolutions to further refine the frequency features and utilize the Inverse Discrete Wavelet Transform (IDWT) to recover the spatial information. Then the recovered spatial information is fused with the features after self-attention. This process can be defined as
\begin{equation}
\small
\begin{aligned}
{x'}_{i} = {\rm DWT}(x_i),  \;
H_i = {\rm Attn}(Q, {x'}_{i} W^{K}_i, {x'}_{i} W^V_i), \;
H'_i = H_i + {\rm IDWT}(\sigma({x'}_{i})),
\end{aligned}
\label{eq:msa}
\end{equation}
where $\sigma(\cdot)$ denotes a convolution layer. 
To increase the efficiency, we reduce the dimension of query and key features to half of the value features.

{\em - Inspective Knowledge In.}
The two branches in our method are highly related, and we believe the knowledge from the inspective branch can provide fine-grained information support for our proposed self-attention module. Considering that the value features in each head represent the content information. We extract the intermediate features from the inspective branch to generate \textbf{\em support value} features, which guides the learning of value features. Denote the corresponding output feature from the inspective branch as $Y$. The value features in $i$-th head can be updated as $x_i W^V_i \psi(Y)$, where $\psi(\cdot)$ denotes operations containing {convolution and global average pooling}. Thus, the self-attention can be finally expressed as $H_i = {\rm Attn}(Q, {x'}_{i} W^{K}_i, {x'}_{i} W^V_i \psi(Y))$

\smallskip
\noindent{\bf Overall operations.}
Let the block index be $\ell$. Given the input feature $X^{\ell}$ at $\ell$-th block. The operations inside this block can be written as
\begin{equation}
\small
\begin{aligned}
 \tilde{X}^{\ell-1} = X^{\ell-1} + {\rm EWSA}({\rm LN}(X^{\ell-1})),   \;
 \tilde{X}^{\ell} = \tilde{X}^{\ell-1} + {\rm FFN}({\rm LN}(\tilde{X}^{\ell-1})),
\end{aligned}
\label{eq:tfb}
\end{equation}
where ${\rm EWSA}$ denotes the proposed efficient wavelet-guided self-attention, ${\rm LN}$ and ${\rm FFN}$ correspond to the LayerNorm~\cite{ioffe2015batch} and the feed-forward network as in~\cite{xie2021segformer,yuan2021hrformer}.

\smallskip
\noindent{\bf Cognitive Knowledge Out.} To promote the learning of fine-grained information, we further fuse the output of this block with intermediate features from the inspective branch, to provide global information priors.

\subsection{Network Architecture} \label{arch}
\smallskip
\noindent{\bf Stem Network.}
We develop a lightweight stem network for extracting task-related features, and these features are then sent to respective branches. In the stem, two convolution layers are first employed. These convolution layers have a kernel size of $3 \times 3$ with stride $2$, which downsample the input image to a size of $1/4$. Then we employ three simple basic residual blocks~\cite{he2016deep}. Each block contains two convolutional layers with a skip connection. The layers in the first blocks have a kernel size of $3 \times 3$ with stride $1$. For the rest two blocks, the layers have the same kernel size with stride $2$ in the first layer. Thus the size of these features after each block becomes $\{ 1/4,1/8,1/16 \}$. The corresponding channel number is $\{ 64,128,128 \}$

\smallskip
\noindent{\bf Cognitive Branch.}
This branch consists of three Efficient Wavelet-guided Transformer blocks. After each block, we perform a convolution to downsample the features. These features are then upsampled and fused with corresponding features from the inspective branch. The size of features after each block is $\{ 1/16,1/32,1/64 \}$, and corresponding channel number is $\{ 128,256,384 \}$.

\smallskip
\noindent{\bf Inspective Branch.}
This branch contains three residual blocks as used in the stem without downsampling. The input feature of this branch is from the second block in the stem. The size of features after each block is $\{ 1/8,1/8,1/8 \}$, and the corresponding channel number is $\{ 128,128,128 \}$. The output of each block is subjected to convolution layers to generate support value for the cognitive branch.

\smallskip
\noindent{\bf Detection Head.} \label{dectection head}
To maintain the efficiency of our method, we employ a simple detection head, which only contains two convolution layers with a kernel size of $3 \times 3$ and $1 \times 1$.

\begin{wrapfigure}{r}{0.35\textwidth}
\centering
\includegraphics[width=0.9\linewidth]{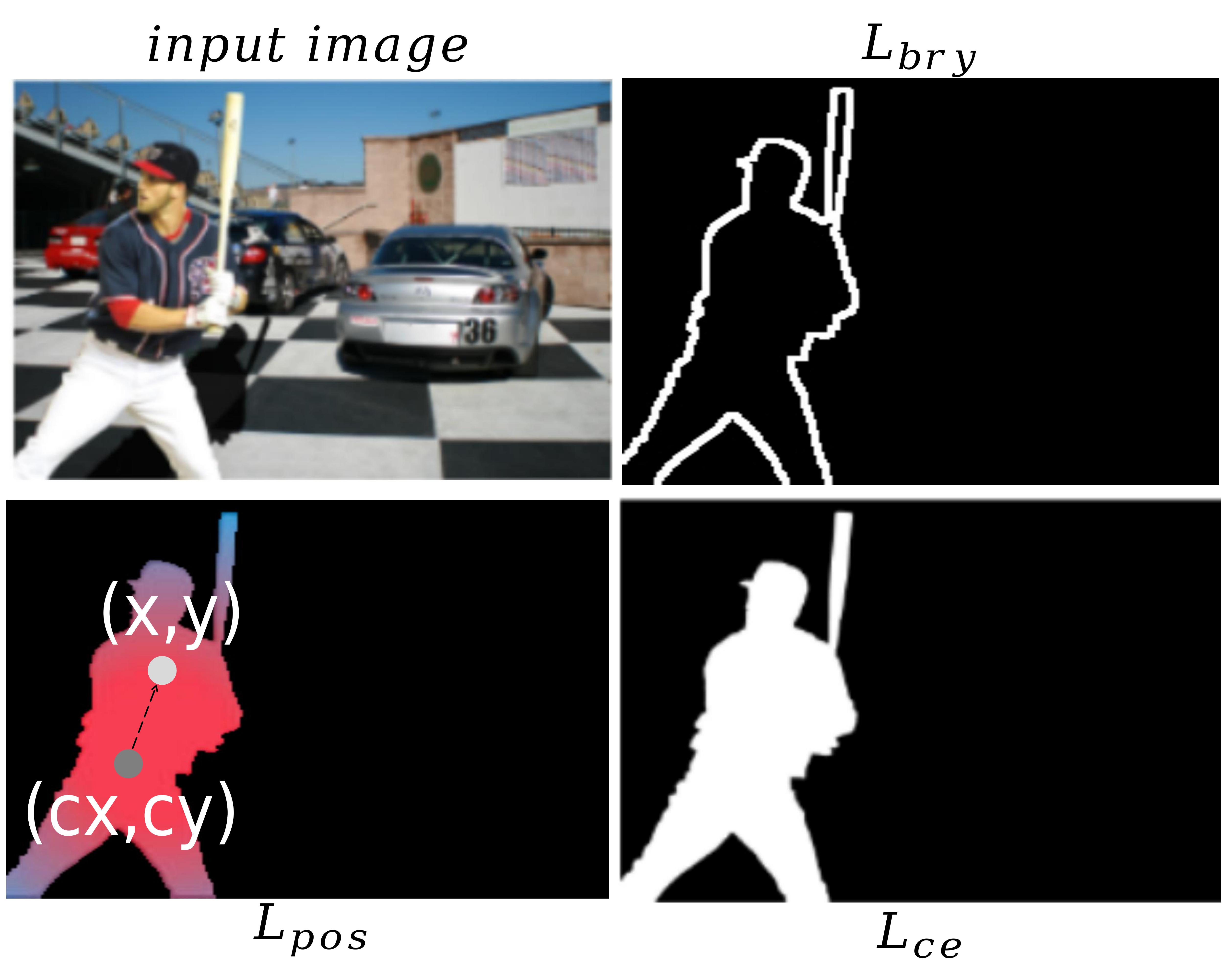} 
\vspace{-0.4cm}
\caption{\small Illustration of each loss term.}
\label{fig:loss}
\end{wrapfigure}

\subsection{Losses}
Our method considers three loss terms, a pixel-wise cross-entropy loss $L_{\rm ce}$, a boundary loss $L_{\rm bry}$, and a position loss $L_{\rm pos}$ (See Fig.~\ref{fig:loss}). The pixel-wise cross-entropy loss $L_{\rm ce}$ is generally employed in this task~\cite{wu2019mantra, liu2022pscc, guo2023hierarchical}, aiming to penalize the deviation of each pixel to the ground truth mask. This loss is performed after detection head. The boundary loss $L_{\rm bry}$ and position loss $L_{\rm pos}$ are auxiliary losses to emphasize the importance of the manipulated boundary and position.

To calculate the boundary loss $L_{\rm bry}$, we utilize the Canny filter~\cite{ding2001canny} on the ground truth mask to extract the boundaries and dilate them with {kernel size of 4}.
Then we design a boundary detection head and employ cross-entropy loss to measure the errors between predicted boundaries and the ground truth boundaries. 

The position loss $L_{\rm pos}$ aims to predict the offset of each manipulated pixel to the center of the manipulated region. Denote $(x,y)$ as the coordinates of pixels in a manipulated region, and $(cx,cy)$ denote the center of this manipulated region. The position loss can be formulated as $L_{\rm pos} = \ell_1(x', (x-cx)) + \ell_1(y', (y-cy))$, where $x',y'$ are predicted offset and employ a $\ell_1$ loss to measure the offset errors between prediction and ground truth.

The prediction head of $L_{\rm bry}$ and $L_{\rm pos}$ is the same as the detection head in $L_{\rm ce}$. The overall objective can be expressed as
$L_{\rm total} = \lambda_1 L_{\rm ce} + \lambda_2 L_{\rm bry} + \lambda_3 L_{\rm pos},$
where $\lambda_1,\lambda_2,\lambda_3$ are weight factors for balancing each objective term.


\section{Experiments} 

\smallskip
\noindent\textbf{Datasets and Evaluation Metrics.} Our method is validated using four widely used datasets, including CASIA~\cite{dong2013casia}, NIST16~\cite{guan2016nist}, Columbia~\cite{ng2009columbia}, and Coverage~\cite{wen2016coverage}. {Following the protocols in previous methods~\cite{chen2018encoder}}, we pre-train the models on synthetic data and fine-tune the models on each dataset. The synthetic data used in our method is from~\cite{liu2022pscc}, and we randomly select a subset of $100,000$ images as the pre-train data for all experiments. 

Following the widely accepted practices, we adopt pixel-level F1 score and Area Under the receiver operating characteristic Curve (AUC) as our evaluation metrics. F1 and AUC measure the binary classification accuracy for every pixel. Both metrics range in $[0, 100]$ percentage, and higher scores indicate better performances. 

\smallskip
\noindent\textbf{Implementation Details.} 
Our method is implemented by PyTorch 1.12~\cite{paszke2019pytorch}. {We follow the evaluation settings for the lightweight segmentation method~\cite{yu2018bisenet,xu2023pidnet} and conduct tests using general GPUs with Nvidia RTX 3060.} In the training phase, we adopt the AdamW optimizer~\cite{loshchilov2018fixing} with a weight decay factor $0.025$. The initial learning rate is set to $10^{-4}$ and decays to $10^{-5}$ by a factor of $10^{-3}$. The image size is set to $224 \times 224$. The batch size is set to $64$. The total iterations is set to $90,000$. The weight factors in the objective are set as $\lambda_1 = 1, \lambda_2 = 2,\lambda_3 = 5$. 

\begin{table*}[!b]
\centering
\small
\caption{\small Performance of FLOPs, Parameters, FPS, F1 score (\%), and AUC (\%).}
\label{table1}
\resizebox{\textwidth}{!}{
\begin{tabular}{l r r r cc cc cc cc cc}
\hline
\multicolumn{1}{l}{\multirow{2}{*}{Method}}  &\multirow{2}{*}{FLOPs $\downarrow$}&\multirow{2}{*}{Params $\downarrow$} &\multirow{2}{*}{FPS $\uparrow$}& \multicolumn{2}{c}{CASIA}&\multicolumn{2}{c}{Columbia}& \multicolumn{2}{c}{NIST16}& \multicolumn{2}{c}{COVERAGE}& \multicolumn{2}{c}{\textbf{Average}}\\
\cline{5-14}

\multicolumn{2}{c}{} & & & $F_1\uparrow$ & AUC$\uparrow$ & $F_1\uparrow$ & AUC$\uparrow$ & $F_1\uparrow$ & AUC$\uparrow$ & $F_1\uparrow$ & AUC$\uparrow$  & $F_1\uparrow$ & AUC$\uparrow$\\
\hline

ManTraNet$^*$~\cite{wu2019mantra} & 191.07G & 3.80M & 16 & --- & 81.7 & --- & 82.4 & --- & 79.5 & --- & 81.9  & --- & 81.4\\

ManTraNet$^\dagger$~\cite{wu2019mantra}  & 191.07G & 3.80M & 16 & 10.5 & 77.2 & 19.1 & 73.3 & 7.1 & 76.3 & 5.4 & 73.7 & 10.5 & 75.1\\

PSCC~\cite{liu2022pscc} &	20.38G & 2.75M & 48 & 44.5 & 80.5 & 16.4 & 52.9 & 68.7 & 98.2 & 33.7 & 76.7 & 40.8 & 77.1 \\

HiFi-Net~\cite{guo2023hierarchical} &	82.3G & 10.2M & 20 & 17.9 & 63.7 & 42.1 & 62.1 & 24.6 & 94.3 & 23.6 & 60.2 & - & - \\

WSCL$^*$~\cite{zhai2023towards} &	$>$4.12G & $>$25.6M & $<$240 & 15.3 & - & 36.2 & - & 9.9 & - & 20.1 & - & 20.4 & - \\

SAFL-Net$^*$~\cite{sun2023safl} &	$>$5.35G & $>$66M & $<$80 & 74.0 & 90.8 & - & 96.9 &  87.9 & 99.7 & 80.3 & 97.0 & 80.7 & 96.1 \\

ObjectFormer$^*$~\cite{wang2022objectformer} & $>$3.16G & $>$38M & $<$120 & 57.9 & 88.2 & - & 95.5 & 82.4 & 99.6 & 75.8 & 95.7 & 72.0 & 94.7 \\

UEN$^*$~\cite{ji2023uncertainty} & $>$34.6G & $>$45.2M & $<$48 & 62.9 & 87.7 & - & - & 93.2 & 99.6 & - & - & 78.0 & 93.6 \\

\hline
BeSiNetV1~\cite{yu2018bisenet} &2.84G	&13.27M & 728 & 33.1 & 77.9& 37.4 & 89.5 & 83.1 & 98.4 & 37.2 & 83.3 & 47.7 & 87.3 \\

BeSiNetV2~\cite{yu2021bisenet}  &2.36G	&3.34M	& 600 & 31.9 & 75.9& 28.4 & 81.8 & 82.1 & 98.0 & 32.9 & 77.3 & 43.8 & 83.3\\


PIDNET~\cite{xu2023pidnet}  &4.29G	&28.53M & 432 & 35.4 & 75.4& 9.1 & 67.6 & 87.9 & 98.4 & 36.3 & 77.8 & 42.2 & 79.8\\

SegNeXt~\cite{guo2022segnext}  & 2.93G	&13.89M & 248 & 9.4 & 68.0& 33.1 & 82.7 & 62.9 & 94.9 & 25.2 & 74.7 & 32.7 & 80.1\\

EfficientViT~\cite{cai2022efficientvit}  &1.73G	&15.27M	& 384 & 35.6& 76.0& 34.5 & 83.0 & 87.0 & 98.8 & 39.7 & 85.5 & 49.2 & 85.8\\

Segformer~\cite{xie2021segformer} & 3.43G	&24.72M & 208 & 7.5 & 72.6& 52.7 & 90.9 & 59.6 & 96.4 & 18.1 & 81.2 & 34.5 & 85.3 \\

DDRNet~\cite{hong2021deep}  & 3.57G &20.30M &  536 & 18.6 & 74.6& 41.6 & 85.9 & 28.8 & 82.3 & 32.5 & 82.1 & 30.4 & 81.2\\
\hline
Ours  &2.16G &8.36M	& 672 & 44.6 & 81.9& 49.3 & 84.7 & 86.5 & 98.9 & 54.7 & 90.6  & 58.8& 89.0 \\
\hline
\end{tabular}}
\label{table1}
\end{table*}

\smallskip
\noindent{\bf Results.} \label{results}
To ensure the fairness of the experimental comparison, we uniformly use our synthetic data to retrain the methods. Specifically, we retrain two dedicated image manipulation detection methods, PSCC~\cite{liu2022pscc} and HiFi-Net~\cite{guo2023hierarchical}, and adapt several efficient semantic segmentation methods into our task, including BiSeNetV1~\cite{yu2018bisenet}, BeSiNetV2~\cite{yu2021bisenet}, DDRNet~\cite{hong2021deep}, SegFormer~\cite{xie2021segformer}, SegNext~\cite{guo2022segnext}, EifficientViT~\cite{cai2022efficientvit} and PIDNet~\cite{xu2023pidnet}. For adaptation, we modify their segmentation head into the manipulation detection head and then pre-trained these methods on our synthetic data. These methods are trained using the same setting as us. 
Note the methods of ObjectFormer~\cite{wang2022objectformer}, WSCL~\cite{zhai2023towards}, SAFL-Net~\cite{sun2023safl}, and UEN~\cite{ji2023uncertainty} have not released their codes (marked by $*$). Thus we could not assess their performance in our setting. Instead, we cite their reported scores and assess the FOLPs, Params and FPS of their backbones as the minimum. Specifically, the backbones of these methods are ResNet50~\cite{he2016deep}, EfficientNetB7~\cite{tan2019efficientnet}, $2 \times$ EfficientB4 and HRNetv2~\cite{wang2020deep}, respectively. ManTraNet~\cite{wu2019mantra} has only released its testing code, thus we directly use its pretrained model for evaluation (marked by $\dagger$). We also include its original report for comparison (marked by $*$).

Table~\ref{table1} shows the performance of FLOPs, Parameters, FPS, F1 score ($\%$), and AUC ($\%$) of different methods on four datasets. Note the $\uparrow$ denotes the larger value represents better performance and vice versa. 
From the results, we can see that our method outperforms the dedicated image manipulation methods by a large margin. For instance, compared to PSCC, we averagely improve $18.0\%$ in the F1 score and $11.9\%$ in the AUC score. Since ManTraNet does not provide the evaluation results in the F1 score, we can only compare us with it in the AUC score. It can also be seen that our method notably improves the AUC score by $7.6\%$ on average. Despite PSCC and ManTraNet having fewer parameters than us, their FLOPs are significantly larger due to the sophisticated forward connects in HRNet and LSTM, leading to a slower inference speed. 
SAFL-Net, ObjectFormer, and UEN are more recent methods that utilize heavy architectures. Although they surpass ours in terms of F1 and AUC, their FLOPs and Params are significantly larger, leading to a much lower FPS than ours. WSCL, a weakly supervised method, exhibits limited performance despite a complex architecture being used.
Compared to the adapted semantic segmentation methods, our method has competitive, even fewer parameters and FLOPs, resulting in a best $672$ FPS. {Moreover, our method surpasses others by $18.6\%$ in F1 score and $6.6\%$ AUC score on average, showing the good balance of our method between performance and efficiency. }

\begin{figure}[!t]
\centering
\includegraphics[width=0.9\linewidth]{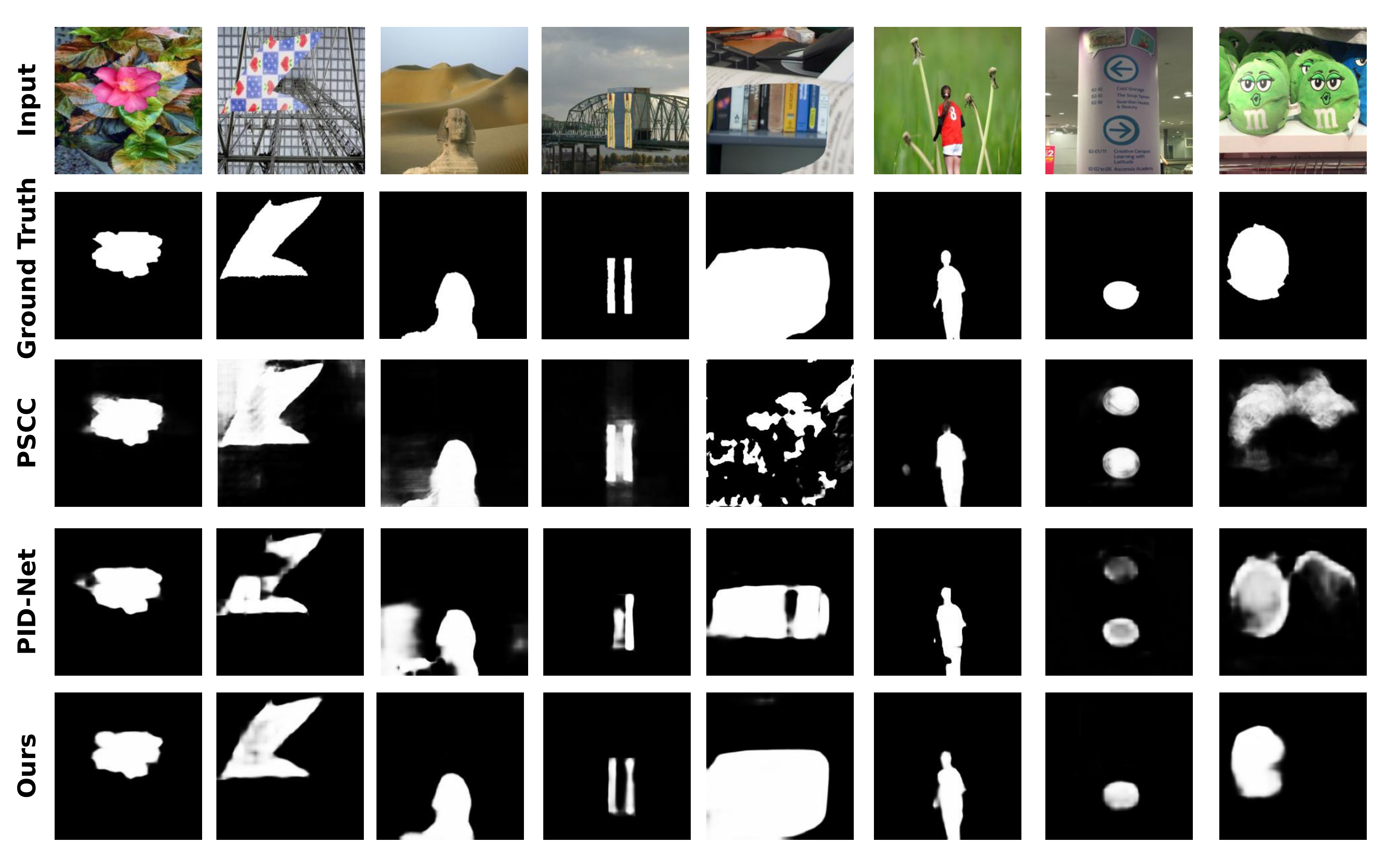} 
\vspace{-0.5cm}
\caption{\small More challenging visual examples.}
\label{fig:5}
\end{figure}


Fig.~\ref{fig:5} shows more detection results of our method on more challenging images. It can be seen that when the manipulation is imperceptive and semantically harmonious, our method can perform better than others to identify these ambiguous regions. 



\smallskip
\noindent{\bf Effect of Each Component.}
Table~\ref{table2} shows the performance of our methods using different components. Denote ``SQ'' as shared global Q, and ``SV'' as support V from the inspective branch. WL denotes the Wavelet module used in IWSA. By adding WL, the performance is improved by $0.8\%$ (F1) and $0.7\%$ in (AUC). The performance is further improved by $0.6\%$ (F1) and $0.3\%$ (AUC) after adding SQ. By using all components, the performance is notably improved by $9.2\%$ (F1) and $3.8\%$ (AUC) on average.

\begin{table}[!t]
    \centering
    \begin{minipage}[c]{0.5\linewidth}
    \caption{\small Effect of components.}
    \setlength{\tabcolsep}{1mm}
    \resizebox{\textwidth}{!}{
    \begin{tabular}{ccccccccccc}
    \hline
    \multirow{2}{*}{WL} & \multirow{2}{*}{SQ} & \multirow{2}{*}{SV} & \multicolumn{2}{c}{CASIA}& \multicolumn{2}{c}{NIST16}& \multicolumn{2}{c}{COVERAGE} & \multicolumn{2}{c}{Average}\\
    \cline{4-11}    
    & & & $F_1\uparrow$ & AUC$\uparrow$ & $F_1\uparrow$ & AUC$\uparrow$ & $F_1\uparrow$ & AUC$\uparrow$ &$F_1\uparrow$ & AUC$\uparrow$ \\
    \hline
    - & - & - & 36.4 & 78.1& 82.8 & 98.7 & 39.0 & 83.2 & 52.7 & 86.6 \\
    \checkmark & - & - & 36.2 & 80.4 & 85.4 &  98.6 & 38.8 & 83.0 & 53.5 & 87.3 \\
    \checkmark & \checkmark & - & 37.6 & 78.1 & 81.4 &  98.7 & 43.4 & 86.0 & 54.1 & 87.6 \\
    \hline
    \checkmark & \checkmark & \checkmark & \textbf{44.6} & \textbf{81.9}& \textbf{86.5} & 98.9 & \textbf{54.7} & \textbf{90.6} & \textbf{61.9} & \textbf{90.4} \\
    \hline
    \end{tabular}}
    \label{table2}
    \end{minipage}
    \hfill
    \begin{minipage}[c]{0.49\linewidth}
    \caption{\small Effect of loss terms.}
    \setlength{\tabcolsep}{1mm}
    \resizebox{\textwidth}{!}{
    \begin{tabular}{ccccccccccc}
    \hline
    \multirow{2}{*}{$L_{\rm{ce}}$} &\multirow{2}{*}{$L_{\rm{bry}}$} & \multirow{2}{*}{$L_{\rm{pos}}$} &\multicolumn{2}{c}{CASIA}& \multicolumn{2}{c}{NIST16}& \multicolumn{2}{c}{COVERAGE} & \multicolumn{2}{c}{Average}\\
    \cline{4-11}    
    & & & $F_1\uparrow$ & AUC$\uparrow$ & $F_1\uparrow$ & AUC$\uparrow$ & $F_1\uparrow$ & AUC$\uparrow$ &$F_1\uparrow$ & AUC$\uparrow$ \\
    \hline    
    \checkmark & - & -	& 37.4 & 79.7& 86.4 & 98.9 & 38.8 & 84.1 & 54.2 & 87.6  \\
    \checkmark & \checkmark & -	& 41.1 & 80.7& 85.5 & 98.7 & 47.8 & 87.6 & 58.1 & 89.0  \\
    \checkmark & - & \checkmark & 39.6 & 81.2& 81.3 & 98.9 & 42.6 & 82.8 & 54.5 & 87.6 \\
    \hline
    \checkmark & \checkmark & \checkmark & \textbf{44.6} & \textbf{81.9}& \textbf{86.5} & \textbf{98.9} & \textbf{54.7} & \textbf{90.6} & \textbf{61.9} & \textbf{90.4} \\
    \hline
    \end{tabular}}
    \label{table3}
    \end{minipage}  
\end{table}

\smallskip
\noindent{\bf Effect of Loss Terms.} Table~\ref{table3} studies the effect of each loss term, showing that without using the boundary and position loss terms, the performance is highly degraded by $7.7\%$ (F1) and $2.8\%$ (AUC) on average. Similarly, without using either boundary loss or position loss can both introduce a performance drop by by $7.4\%$ (F1) and $2.8\%$ (AUC), and $3.8\%$ (F1) and $1.4\%$ (AUC) on average. These results demonstrate the importance of predicting the manipulation boundaries and positions.

    
    

\begin{table}[!t]
    \centering
    \begin{minipage}[c]{0.6\linewidth}
    \small
    \centering
    \vspace{0.2cm}
    \caption{\small Effect of EWTB.}
    \setlength{\tabcolsep}{1mm}
    \resizebox{\textwidth}{!}{
    \begin{tabular}{lcccccccc}
    \hline
    \multicolumn{1}{l}{\multirow{2}{*}{Method}}& \multicolumn{2}{c}{CASIA}& \multicolumn{2}{c}{NIST16}& \multicolumn{2}{c}{COVERAGE} & \multicolumn{1}{l}{\multirow{2}{*}{FLOPs}} & \multicolumn{1}{l}{\multirow{2}{*}{Params}}\\
    \cline{2-7}
    
    \multicolumn{1}{c}{} & $F_1\uparrow$ & AUC$\uparrow$ & $F_1\uparrow$ & AUC$\uparrow$ & $F_1\uparrow$ & AUC$\uparrow$  \\
    \hline
    
    Vanilla 	    & 31.6 & 75.4 & 72.2 & 97.9 & 37.5 & 81.6  & 6.23G & 24.21M \\
    EDTB             & 40.1 & 79.8 & 82.8 & 98.5 & 40.8 & 83.6 & 2.16G & 8.36M \\ 
    EWTB & \textbf{44.6} & \textbf{81.9}& \textbf{86.5} & \textbf{98.9} & \textbf{54.7} & \textbf{90.6} &\textbf{2.16G} &\textbf{8.36M} \\
    \hline
    \end{tabular}}
    \label{vanllia}
    \end{minipage}
    \hfill
    \begin{minipage}[c]{0.39\linewidth}
    \small
    \centering  
    \caption{\small Various number of stages.} 
    \setlength{\tabcolsep}{1mm}
    \resizebox{0.8\textwidth}{!}{ 
    \begin{tabular}{c |cc |cc}  
        \hline
    \multicolumn{1}{c|}{Stage} & $F_1\uparrow$  & AUC$\uparrow$  & Params & FLOPs \\ \hline
    2                      & 80.7 & 98.5 & 3.85M   & 1.86G  \\
    3                      & \textbf{86.5} & \textbf{98.9} & 8.37M   & 2.16G \\  
    4                      & 81.3 & 98.5 & 10.71M  & 2.43G  \\ \hline 
    \end{tabular}  
    \label{table6}}
    \end{minipage}
    \vspace{-0.5cm}
\end{table}

\smallskip
\noindent{\bf Effect of EWTB.}
To better validate the effectiveness of EWTB, we adapt vanilla Transformer blocks (output channel dimension is $768$) and DCT into our architecture. Table~\ref{vanllia} shows the compared results. It can be seen that our method outperforms the vanilla version by a large margin while having fewer FLOPs and Params. Compared to integrating DCT, our method shows better performance, demonstrating the efficacy of using Wavelet transformation. 

\smallskip
\noindent\textbf{Various Numbers of Stages.} This part studies the effect of stage numbers from 2 to 4. The results in table~\ref{table6} show that extending or reducing the number of stages can slightly affect the performance. This is likely because the stage number is highly related to the role of the cognitive and inspective branches. When the stage is increased, the model may overfit to a certain distribution. On the other hand, models with fewer stages may not adequately learn the key features required to identify manipulation traces.



\begin{figure}[!t]
    \centering
    \includegraphics[width=\linewidth]{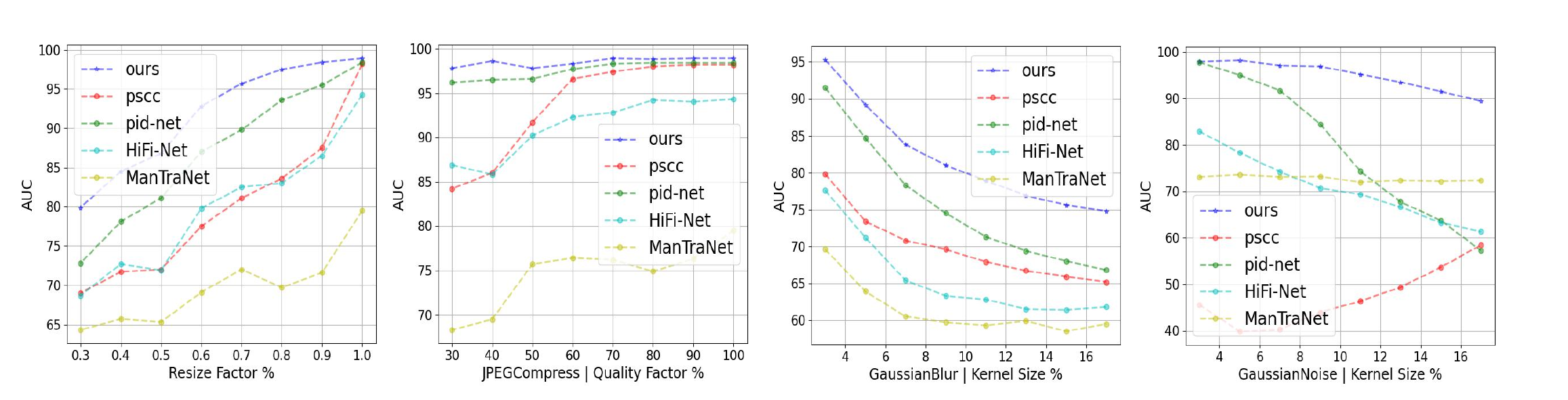} 
    \vspace{-0.4cm}
    \caption{\small Robustness analysis of different methods.}
    \label{fig:7}
\end{figure}

\smallskip
\noindent{\bf Robustness Analysis.}
To study the robustness, we perform various preprocessing operations on the images, including applying GaussianBlur, performing JPEG compression (JPEGCompress), adding GaussianNoise, and resizing the images (Resize). As can be seen from Fig.~\ref{fig:7}, our method performs better than others, showing a certain robustness ability against these operations. When confronting the operations of GaussianBlur and GaussianNoise, our method drops notably compared to JPEGCompress and Resize. We conjecture that the manipulation traces are highly reflected around boundaries, and these two operations may largely disturb the image structural information, thus affecting the final results.

\section{Conclusion}
This paper describes an efficient two-stream design dedicated to real-time image manipulation detection. The architecture comprises a cognitive branch and an inspective branch, designed to capture global and local traces. In the cognitive branch, we introduce the Efficient Wavelet-guided Transformer Block (EWTB) to extract manipulation traces related to frequency from a global view. This block interacts with the inspective branch, aiming to mutually enhance both streams. Experimental results demonstrate the efficacy and efficiency of our method in detecting image manipulations.

\bigskip
\noindent\textbf{Acknowledgement: }
 This work was supported in part by the National Natural Science Foundation of China under Grant No.62402464, China Postdoctoral Science Foundation under Grant No.2021TQ0314 and Grant No.2021M703036.

\bibliography{egbib}

\begin{thebibliography}{38}
\providecommand{\natexlab}[1]{#1}
\providecommand{\url}[1]{\texttt{#1}}
\expandafter\ifx\csname urlstyle\endcsname\relax
  \providecommand{\doi}[1]{doi: #1}\else
  \providecommand{\doi}{doi: \begingroup \urlstyle{rm}\Url}\fi

\bibitem[Cai et~al.(2022)Cai, Gan, and Han]{cai2022efficientvit}
Han Cai, Chuang Gan, and Song Han.
\newblock Efficientvit: Enhanced linear attention for high-resolution low-computation visual recognition.
\newblock \emph{arXiv preprint arXiv:2205.14756}, 2022.

\bibitem[Chen et~al.(2018)Chen, Zhu, Papandreou, Schroff, and Adam]{chen2018encoder}
Liang-Chieh Chen, Yukun Zhu, George Papandreou, Florian Schroff, and Hartwig Adam.
\newblock Encoder-decoder with atrous separable convolution for semantic image segmentation.
\newblock In \emph{Proceedings of the European conference on computer vision (ECCV)}, pages 801--818, 2018.

\bibitem[Ding and Goshtasby(2001)]{ding2001canny}
Lijun Ding and Ardeshir Goshtasby.
\newblock On the canny edge detector.
\newblock \emph{Pattern recognition}, 34\penalty0 (3):\penalty0 721--725, 2001.

\bibitem[Dong et~al.(2013)Dong, Wang, and Tan]{dong2013casia}
Jing Dong, Wei Wang, and Tieniu Tan.
\newblock Casia image tampering detection evaluation database.
\newblock In \emph{2013 IEEE China summit and international conference on signal and information processing}. IEEE, 2013.

\bibitem[Guan et~al.(2016)Guan, Lee, Yates, Delgado, Zhou, Joy, and Pereira]{guan2016nist}
HY~Guan, YY~Lee, A~Yates, A~Delgado, D~Zhou, D~Joy, and A~Pereira.
\newblock Nist nimble 2016 datasets, 2016.

\bibitem[Guo et~al.(2022)Guo, Lu, Hou, Liu, Cheng, and Hu]{guo2022segnext}
Meng-Hao Guo, Cheng-Ze Lu, Qibin Hou, Zhengning Liu, Ming-Ming Cheng, and Shi-Min Hu.
\newblock Segnext: Rethinking convolutional attention design for semantic segmentation.
\newblock \emph{Advances in Neural Information Processing Systems}, 35:\penalty0 1140--1156, 2022.

\bibitem[Guo et~al.(2023)Guo, Liu, Ren, Grosz, Masi, and Liu]{guo2023hierarchical}
Xiao Guo, Xiaohong Liu, Zhiyuan Ren, Steven Grosz, Iacopo Masi, and Xiaoming Liu.
\newblock Hierarchical fine-grained image forgery detection and localization.
\newblock In \emph{Proceedings of the IEEE/CVF Conference on Computer Vision and Pattern Recognition}, pages 3155--3165, 2023.

\bibitem[He et~al.(2016)He, Zhang, Ren, and Sun]{he2016deep}
Kaiming He, Xiangyu Zhang, Shaoqing Ren, and Jian Sun.
\newblock Deep residual learning for image recognition.
\newblock In \emph{Proceedings of the IEEE conference on computer vision and pattern recognition}, pages 770--778, 2016.

\bibitem[Hong et~al.(2021)Hong, Pan, Sun, and Jia]{hong2021deep}
Yuanduo Hong, Huihui Pan, Weichao Sun, and Yisong Jia.
\newblock Deep dual-resolution networks for real-time and accurate semantic segmentation of road scenes.
\newblock \emph{arXiv preprint arXiv:2101.06085}, 2021.

\bibitem[Hu et~al.(2020)Hu, Zhang, Jiang, Chaudhuri, Yang, and Nevatia]{hu2020span}
Xuefeng Hu, Zhihan Zhang, Zhenye Jiang, Syomantak Chaudhuri, Zhenheng Yang, and Ram Nevatia.
\newblock Span: Spatial pyramid attention network for image manipulation localization.
\newblock In \emph{Computer Vision--ECCV 2020: 16th European Conference, Glasgow, UK, August 23--28, 2020, Proceedings, Part XXI 16}, pages 312--328. Springer, 2020.

\bibitem[Ioffe and Szegedy(2015)]{ioffe2015batch}
Sergey Ioffe and Christian Szegedy.
\newblock Batch normalization: Accelerating deep network training by reducing internal covariate shift.
\newblock In \emph{International conference on machine learning}, pages 448--456. pmlr, 2015.

\bibitem[Ji et~al.(2023)Ji, Chen, Guo, Xu, Wang, and Chen]{ji2023uncertainty}
Kaixiang Ji, Feng Chen, Xin Guo, Yadong Xu, Jian Wang, and Jingdong Chen.
\newblock Uncertainty-guided learning for improving image manipulation detection.
\newblock In \emph{Proceedings of the IEEE/CVF International Conference on Computer Vision}, pages 22456--22465, 2023.

\bibitem[Li et~al.(2020)Li, Shen, Guo, and Lai]{li2020wavelet}
Qiufu Li, Linlin Shen, Sheng Guo, and Zhihui Lai.
\newblock Wavelet integrated cnns for noise-robust image classification.
\newblock In \emph{Proceedings of the IEEE/CVF conference on computer vision and pattern recognition}, pages 7245--7254, 2020.

\bibitem[Liu et~al.(2023{\natexlab{a}})Liu, Tan, Tan, Wei, Zhao, and Wang]{liu2023forgery}
Huan Liu, Zichang Tan, Chuangchuang Tan, Yunchao Wei, Yao Zhao, and Jingdong Wang.
\newblock Forgery-aware adaptive transformer for generalizable synthetic image detection.
\newblock \emph{arXiv preprint arXiv:2312.16649}, 2023{\natexlab{a}}.

\bibitem[Liu et~al.(2022)Liu, Liu, Chen, and Liu]{liu2022pscc}
Xiaohong Liu, Yaojie Liu, Jun Chen, and Xiaoming Liu.
\newblock Pscc-net: Progressive spatio-channel correlation network for image manipulation detection and localization.
\newblock \emph{IEEE Transactions on Circuits and Systems for Video Technology}, 2022.

\bibitem[Liu et~al.(2023{\natexlab{b}})Liu, Peng, Zheng, Yang, Hu, and Yuan]{liu2023efficientvit}
Xinyu Liu, Houwen Peng, Ningxin Zheng, Yuqing Yang, Han Hu, and Yixuan Yuan.
\newblock Efficientvit: Memory efficient vision transformer with cascaded group attention.
\newblock In \emph{Proceedings of the IEEE/CVF Conference on Computer Vision and Pattern Recognition}, pages 14420--14430, 2023{\natexlab{b}}.

\bibitem[Liu et~al.(2018)Liu, Sun, Zhou, Huang, and Darrell]{liu2018rethinking}
Zhuang Liu, Mingjie Sun, Tinghui Zhou, Gao Huang, and Trevor Darrell.
\newblock Rethinking the value of network pruning.
\newblock \emph{arXiv preprint arXiv:1810.05270}, 2018.

\bibitem[Loshchilov and Hutter(2018)]{loshchilov2018fixing}
Ilya Loshchilov and Frank Hutter.
\newblock Fixing weight decay regularization in adam.
\newblock 2018.

\bibitem[Ng et~al.(2009)Ng, Hsu, and Chang]{ng2009columbia}
Tian-Tsong Ng, Jessie Hsu, and Shih-Fu Chang.
\newblock Columbia image splicing detection evaluation dataset.
\newblock \emph{DVMM lab. Columbia Univ CalPhotos Digit Libr}, 2009.

\bibitem[Paszke et~al.(2019)Paszke, Gross, Massa, Lerer, Bradbury, Chanan, Killeen, Lin, Gimelshein, Antiga, et~al.]{paszke2019pytorch}
Adam Paszke, Sam Gross, Francisco Massa, Adam Lerer, James Bradbury, Gregory Chanan, Trevor Killeen, Zeming Lin, Natalia Gimelshein, Luca Antiga, et~al.
\newblock Pytorch: An imperative style, high-performance deep learning library.
\newblock \emph{Advances in neural information processing systems}, 32, 2019.

\bibitem[Sun et~al.(2023)Sun, Jiang, Wang, Li, and Cao]{sun2023safl}
Zhihao Sun, Haoran Jiang, Danding Wang, Xirong Li, and Juan Cao.
\newblock Safl-net: Semantic-agnostic feature learning network with auxiliary plugins for image manipulation detection.
\newblock In \emph{Proceedings of the IEEE/CVF International Conference on Computer Vision}, pages 22424--22433, 2023.

\bibitem[Tan and Le(2019)]{tan2019efficientnet}
Mingxing Tan and Quoc Le.
\newblock Efficientnet: Rethinking model scaling for convolutional neural networks.
\newblock In \emph{International conference on machine learning}, pages 6105--6114. PMLR, 2019.

\bibitem[Vaswani et~al.(2017)Vaswani, Shazeer, Parmar, Uszkoreit, Jones, Gomez, Kaiser, and Polosukhin]{vaswani2017attention}
Ashish Vaswani, Noam Shazeer, Niki Parmar, Jakob Uszkoreit, Llion Jones, Aidan~N Gomez, {\L}ukasz Kaiser, and Illia Polosukhin.
\newblock Attention is all you need.
\newblock \emph{Advances in neural information processing systems}, 30, 2017.

\bibitem[Wang et~al.(2020)Wang, Sun, Cheng, Jiang, Deng, Zhao, Liu, Mu, Tan, Wang, et~al.]{wang2020deep}
Jingdong Wang, Ke~Sun, Tianheng Cheng, Borui Jiang, Chaorui Deng, Yang Zhao, Dong Liu, Yadong Mu, Mingkui Tan, Xinggang Wang, et~al.
\newblock Deep high-resolution representation learning for visual recognition.
\newblock \emph{IEEE transactions on pattern analysis and machine intelligence}, 43\penalty0 (10):\penalty0 3349--3364, 2020.

\bibitem[Wang et~al.(2022)Wang, Wu, Chen, Han, Shrivastava, Lim, and Jiang]{wang2022objectformer}
Junke Wang, Zuxuan Wu, Jingjing Chen, Xintong Han, Abhinav Shrivastava, Ser-Nam Lim, and Yu-Gang Jiang.
\newblock Objectformer for image manipulation detection and localization.
\newblock In \emph{Proceedings of the IEEE/CVF Conference on Computer Vision and Pattern Recognition}, pages 2364--2373, 2022.

\bibitem[Wen et~al.(2016)Wen, Zhu, Subramanian, Ng, Shen, and Winkler]{wen2016coverage}
Bihan Wen, Ye~Zhu, Ramanathan Subramanian, Tian-Tsong Ng, Xuanjing Shen, and Stefan Winkler.
\newblock Coverage—a novel database for copy-move forgery detection.
\newblock In \emph{2016 IEEE international conference on image processing (ICIP)}. IEEE, 2016.

\bibitem[Wu et~al.(2019)Wu, AbdAlmageed, and Natarajan]{wu2019mantra}
Yue Wu, Wael AbdAlmageed, and Premkumar Natarajan.
\newblock Mantra-net: Manipulation tracing network for detection and localization of image forgeries with anomalous features.
\newblock In \emph{Proceedings of the IEEE/CVF Conference on Computer Vision and Pattern Recognition}, 2019.

\bibitem[Xie et~al.(2021)Xie, Wang, Yu, Anandkumar, Alvarez, and Luo]{xie2021segformer}
Enze Xie, Wenhai Wang, Zhiding Yu, Anima Anandkumar, Jose~M Alvarez, and Ping Luo.
\newblock Segformer: Simple and efficient design for semantic segmentation with transformers.
\newblock \emph{Advances in Neural Information Processing Systems}, 34:\penalty0 12077--12090, 2021.

\bibitem[Xu et~al.(2023)Xu, Xiong, and Bhattacharyya]{xu2023pidnet}
Jiacong Xu, Zixiang Xiong, and Shankar~P Bhattacharyya.
\newblock Pidnet: A real-time semantic segmentation network inspired by pid controllers.
\newblock In \emph{Proceedings of the IEEE/CVF Conference on Computer Vision and Pattern Recognition}, 2023.

\bibitem[Yang et~al.(2021)Yang, Yin, Molchanov, Li, and Kautz]{yang2021nvit}
Huanrui Yang, Hongxu Yin, Pavlo Molchanov, Hai Li, and Jan Kautz.
\newblock Nvit: Vision transformer compression and parameter redistribution.
\newblock 2021.

\bibitem[Yao et~al.(2022)Yao, Pan, Li, Ngo, and Mei]{yao2022wave}
Ting Yao, Yingwei Pan, Yehao Li, Chong-Wah Ngo, and Tao Mei.
\newblock Wave-vit: Unifying wavelet and transformers for visual representation learning.
\newblock In \emph{European Conference on Computer Vision}, pages 328--345. Springer, 2022.

\bibitem[Yu et~al.(2018)Yu, Wang, Peng, Gao, Yu, and Sang]{yu2018bisenet}
Changqian Yu, Jingbo Wang, Chao Peng, Changxin Gao, Gang Yu, and Nong Sang.
\newblock Bisenet: Bilateral segmentation network for real-time semantic segmentation.
\newblock In \emph{Proceedings of the European conference on computer vision (ECCV)}, 2018.

\bibitem[Yu et~al.(2021)Yu, Gao, Wang, Yu, Shen, and Sang]{yu2021bisenet}
Changqian Yu, Changxin Gao, Jingbo Wang, Gang Yu, Chunhua Shen, and Nong Sang.
\newblock Bisenet v2: Bilateral network with guided aggregation for real-time semantic segmentation.
\newblock \emph{International Journal of Computer Vision}, 129:\penalty0 3051--3068, 2021.

\bibitem[Yuan et~al.(2021)Yuan, Fu, Huang, Lin, Zhang, Chen, and Wang]{yuan2021hrformer}
Yuhui Yuan, Rao Fu, Lang Huang, Weihong Lin, Chao Zhang, Xilin Chen, and Jingdong Wang.
\newblock Hrformer: High-resolution vision transformer for dense predict.
\newblock \emph{Advances in Neural Information Processing Systems}, 34:\penalty0 7281--7293, 2021.

\bibitem[Zampoglou et~al.(2017)Zampoglou, Papadopoulos, and Kompatsiaris]{zampoglou2017large}
Markos Zampoglou, Symeon Papadopoulos, and Yiannis Kompatsiaris.
\newblock Large-scale evaluation of splicing localization algorithms for web images.
\newblock \emph{Multimedia Tools and Applications}, 2017.

\bibitem[Zhai et~al.(2023)Zhai, Luan, Doermann, and Yuan]{zhai2023towards}
Yuanhao Zhai, Tianyu Luan, David Doermann, and Junsong Yuan.
\newblock Towards generic image manipulation detection with weakly-supervised self-consistency learning.
\newblock In \emph{Proceedings of the IEEE/CVF International Conference on Computer Vision}, pages 22390--22400, 2023.

\bibitem[Zhou et~al.(2023)Zhou, Ma, Du, Alhammadi, and Feng]{zhou2023pre}
Jizhe Zhou, Xiaochen Ma, Xia Du, Ahmed~Y Alhammadi, and Wentao Feng.
\newblock Pre-training-free image manipulation localization through non-mutually exclusive contrastive learning.
\newblock In \emph{Proceedings of the IEEE/CVF International Conference on Computer Vision}, pages 22346--22356, 2023.

\bibitem[Zhou et~al.(2018)Zhou, Han, Morariu, and Davis]{zhou2018learning}
Peng Zhou, Xintong Han, Vlad~I Morariu, and Larry~S Davis.
\newblock Learning rich features for image manipulation detection.
\newblock In \emph{Proceedings of the IEEE conference on computer vision and pattern recognition}, pages 1053--1061, 2018.

\end{thebibliography}
\end{document}